\documentclass{article}

\usepackage{arxiv}

\usepackage[utf8]{inputenc} 
\usepackage[T1]{fontenc}    
\usepackage{hyperref}       
\usepackage{url}            
\usepackage{booktabs}       
\usepackage{amsfonts}       
\usepackage{nicefrac}       
\usepackage{microtype}      
\usepackage{lipsum}
\usepackage{subcaption}
\usepackage{graphicx}
\graphicspath{ {./images/} }
\usepackage{natbib}
\usepackage{amssymb}
\usepackage{pifont}
\usepackage{amsmath}
\usepackage{fvextra}
\fvset{breaklines=true, breakanywhere=true, fontsize=\footnotesize}

\title{One Run Is Not Enough: A Monte Carlo Protocol for Evaluating LLM Reliability on Data Science Tasks}

\author{
Wenqi Ni \thanks{Corresponding email: wenqi.ni@ed.ac.uk}\\
University of Edinburgh \\
\texttt{wenqi.ni@ed.ac.uk}\\
\And
Ozan Evkaya\\
University of Edinburgh \\
\texttt{Ozan.Evkaya@ed.ac.uk}\\
\And
Miguel de Carvalho\\
University of Edinburgh \\
\texttt{Miguel.deCarvalho@ed.ac.uk}
}

\begin{document}
\maketitle

\begin{abstract}
Large language models (LLMs) have shown promise for automating data science workflows. Benchmarks such as DataSciBench (2026) and StatQA (2024) have advanced the evaluation of their capabilities across data science and statistical tasks. Assessing their reliability requires examining both the quality of their responses and their consistency across repeated runs, task types, and difficulty levels. With this aim, we introduce DS-QA, a benchmark comprising 100 data science and statistics questions drawn from UK university courses, spanning 16 datasets and three difficulty levels. We develop a three-layer evaluation framework that combines general response properties (runtime, verbosity and text similarity), coarse-grained metrics (accuracy, code executability and completeness), and task-specific metrics. We evaluate OpenAI models using a Monte Carlo protocol with 100 repeated runs. The results reveal substantial variability in responses to data visualisation and supervised learning tasks, particularly at medium and hard difficulty levels, highlighting the importance of evaluating consistency alongside response quality. We also find that an increase of 0.1 in the word verbosity ratio corresponds to an increase of about 0.013 in the token verbosity ratio.

\end{abstract}

\keywords{Large Language Models, LLM Evaluation, Monte Carlo Study, Data Science, Benchmark}

\begin{figure*}[!ht]
\centering
\includegraphics[width=.9\linewidth]{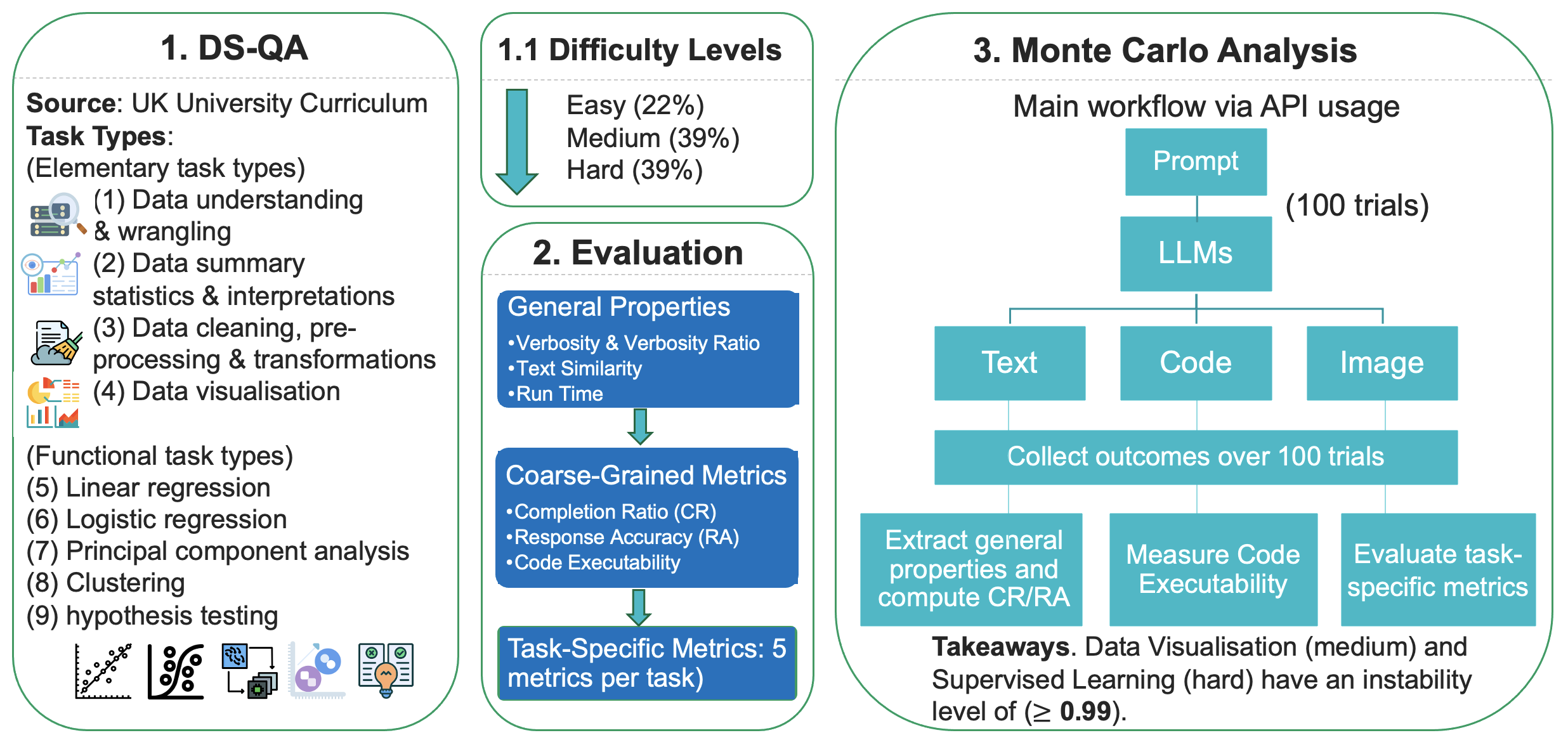}
\caption{The proposed \textsc{DS-QA} contains three key components (cf. Sections \ref{sec:ds-qa} and \ref{sec:evalframework}). (1) A university-level data science benchmark, which covers 9 task types and 3 difficulty levels. (2) A multilayer evaluation framework, incorporating 4 general properties, 3 coarse-grained metrics, and in total 35 task-specific metrics. (3). A Monte Carlo analysis is used to study the reliability of LLM (text, code and image) outputs.
  }
\label{fig:teaser}

\end{figure*}

\section{Introduction}
The recent development of large language models (LLMs) has demonstrated impressive capabilities across a range of scientific tasks including mathematics, physics, and chemistry. Despite their successes, the effectiveness of LLMs in handling complex statistical tasks remains systematically under-explored. While ChatGPT’s Advanced Data Analysis (ADA) can often match or even outperform human analysts in basic tasks, it may struggle with contextually complex or nuanced analyses \citep{Ping2020}. LLMs are unable to make careful assumptions and claims when providing insights, unlike professional human data analysts \citep{Bayne2015,Zhu2024,jing2025dsbench}. Moreover, the inherited randomness of these LLMs can add a certain level of uncertainty to the created responses even if the same prompt is asked repeatedly \citep{evkaya2026using}. 

To fill this gap, our work aims to apply a Monte Carlo assessment to measure the stability of LLM-generated responses. Each question or request is tested 100 times under an identical configuration and then evaluated with the use of our proposed evaluation framework.

\textbf{Contributions.} 
\begin{itemize}
    \item \textbf{A university-level data science question-answer benchmark:} \textsc{DS-QA} includes 100 data science question-answer pairs over 16 different datasets across 9 question dimensions and 3 difficulty levels. The questions were mainly extracted from the UK university curriculum. The question dimensions mainly include: 1. Data Understanding and Wrangling (14\%), 2. Data Cleaning, Pre-processing and Transformations (5\%), 3. Data Summary Statistics and Interpretations (15\%), 4. Data visualisation (19\%), 5. Statistical Models (44\%), and 6. Statistical Hypothesis Testing (3\%). \textsc{DS-QA} enriches the task type of Statistical Models to be Linear Regression, Logistic Regression, Principal Component Analysis, and Clustering. See Figure \ref{fig:conceptvsdifficulty} for the difficulty level distribution. 

    \item \textbf{A multi-layer evaluation framework for common data science tasks:} The framework consists of general properties, coarse-grained, and task-specific metrics. See Figure \ref{fig:chart_evaluate} for an overview. 

    \item \textbf{A Monte Carlo assessment of response reliability:} We examine the response reliability of LLMs across repeated runs per task type and per difficulty level. The key findings are: (1) LLMs show extremely high response deviations when facing Data Visualisation and Supervised Learning tasks, especially at medium and hard difficulty levels. Refer to Figure \ref{fig:result.overall_std}. (2) Word verbosity ratio has a measurable effect on token verbosity ratio: an increase of 0.1 in the word verbosity ratio brings an increase of about 0.013 in the token verbosity ratio (see Figure \ref{fig:result.verbosity}), which helps to evaluate LLM redundancy length from the perspective of computational or API costs. (3) \textsc{gpt-4o-mini} shows the highest level of text similarity with reasonable token usage, compared to \textsc{gpt-4o} and \textsc{gpt-4.1-nano}. (4) For \textsc{gpt-4.1-mini}, a temperature larger than 1 is more likely to bring random changes to responses. These findings have direct implications for researchers deploying LLMs in data science pipelines. 
    
\end{itemize}

The benchmark and code are available at: \url{https://github.com/niwq15/LLM-DS-Reliability.git}.

\section{Related Literature}
\label{sec:literature}
\begin{table*}[ht]
\centering
\begin{tabular}{@{}p{0.15\linewidth} p{0.15\linewidth}  p{0.32\linewidth} p{0.17\linewidth} p{0.12\linewidth}@{}}
\toprule
\textbf{Benchmark} 
&\textbf{Problem Origin} &\textbf{Task Coverage}
&\textbf{Evaluation Metrics}
&\textbf{Consistency Analysis} \\
\midrule
MLAgentBench \citep{Huang2023} 
& Kaggle & 5 question types: Classification, Regression, Segmentation, Language Modeling, Code Speed Improvement.
& Acc.\ + Success Rate, Human &\ding{55} \\

LiveCodeBench \citep{Jain2024} 
&LeetCode \& AtCoder \& CodeForces 
&Focus on Competitive Programming and Code Generation
& Test Cases + Pass Rate &\ding{55} \\

DS-1000 \citep{Lai2023} & StackOverflow 
&Focus on Programming and Code Generation
& Test Cases + Surface‐Form Constraints &\ding{55} \\

AIRepr \citep{zeng2025} 
& DiscoveryBench \& QRData \& StatQA & 
6 question types: Quantitative Reasoning, Statistical Analysis from \citep{Zhu2024}.
& Reproducibility &\ding{55} \\

\hline
Text2Analysis \citep{He2023}
& Human\&LLM & Tabular tasks
&Executable Code Ratio, Acc., Regression Scores &\ding{55} \\ 

InfiAgent-DABench \citep{Hu2024} 
& LLM  & Tabular tasks
& Accuracy &\ding{55} \\ \hline

StatQA \citep{Zhu2024}
& Human
& 5 question types: Descriptive statistics, Correlation Analysis, Contingency Table Tests, Distribution Compliance Tests, and Variance Tests 
&Accuracy and Reproducibility &\ding{55} \\

U-MATH \citep{chernyshev-etal-2025-u} 
& Human (University Curriculum)  
& 6 subjects: Algebra, Differential Calculus, Integral Calculus, Multivariable Calculus, Precalculus, and Sequences and Series.
& Accuracy &\ding{55} \\ 

DataSciBench \citep{Zhang2025} 
& CodeGeeX \& BCB \& Human 
&5 question types: Data Cleaning, Data Understanding, Visualisation, Predictive Modelling, Data Mining and Pattern Recognition, Interpretability and Report generation.
&25 Aggregate Metrics and Programmatic Rules 
&\ding{55} \\ 
\hline
\textsc{DS-QA (Ours)} 
& \textbf{Human (University Curriculum)}
& \textbf{16 datasets, 9 question types, 3 difficulty levels; new: Logistic Regression}
&\textbf{3-layer framework with 42 metrics}  &\ding{51}(\textbf{100 trials}) \\
\bottomrule
\end{tabular}
\caption{Comparison with related work across key dimensions}
\label{tab:comparison}
\end{table*}

\textbf{LLMs on Data Science.}
The systematic evaluation of LLM has attracted attention due to the importance of ensuring the reliability of AI responses. Early benchmarks on mathematics such as FrontierMath \cite{glazer2024frontiermath} and IMProofBench \cite{schmitt2025improofbench} established the foundations of LLM evaluation on expert-level mathematics, moving the focus from final-answer questions to research-level questions requiring a detailed proof. More recent work has focused on data science workflows, such as DS-1000 \citep{Lai2023}, DataSciBench \citep{Zhang2025}, and U-MATH \cite{chernyshev-etal-2025-u}. DS-1000 \cite{Lai2023} proposes a benchmark of 1,000 programming and coding generation problems from StackOverflow. U-MATH \cite{chernyshev-etal-2025-u} introduces a benchmark of 1,100 university-level problems from current US curriculum, with 20\% of problems involving visual elements. DataSciBench \citep{Zhang2025} proposes a Task–Function–Code (TFC) framework across 6 data science task types with precisely defined metrics and programmatic rules. They also collect prompts from CodeGeeX, BigCodeBench (BCB), LLM-synthesis and human experts.

However, the existing benchmarks evaluate performance mostly under single-pass prompting without taking response variability and consistency into account; and the main problem sources, except U-MATH \cite{chernyshev-etal-2025-u}, are platform crawling or LLM synthesis, lacking academic stratification based on real university courses. Our DS-QA benchmark comes from a university-level statistics curriculum and covers 9 common data-science task types with three difficulty levels, filling in the above gaps.

\textbf{Data Science Task Automation.}
An influential theoretical framework that characterises the data science (DS) task categories is De Bie et al. \cite{DeBie2022}. The work introduces four quadrants of DS, which are Data Engineering, Data Exploration, Model Building, and Exploitation. On this basis, subsequent work has enriched the DS task types and examined their LLM-assisted implementation, such as exploratory data analysis (e.g. StatQA \cite{Zhu2024} and DataSciBench \cite{Zhang2025}), code generation for statistical modelling (e.g. DS-1000 \cite{Lai2023} and MLAgentBench \citep{Huang2023}). LLM tools such as OpenAI's Advanced Data Analytics (ADA) have extended these capabilities and enabled end-to-end DS task completion.

Despite this development, systematic evaluation of these tools under realistic DS conditions remains underdeveloped. Existing studies tend to assess task completion on individual problems without enough attention to explore whether performance keeps consistent across iterations or how model behaviour varies by task types and difficulty levels. The absence of a principled evaluation framework also makes it hard for practitioners to compare across benchmarks on the same task types and in the meanwhile limit actionable guidance on how prompting structuring can help to mitigate failure modes. 

By adopting the four-quadrant DS taxonomy from \cite{DeBie2022} as the basis, our work fills the gaps mentioned above by providing both a structured question-answer benchmark and a three-layer evaluation framework (including general properties, coarse-grained metrics, and task-specific metrics) in which the criteria also serve as guidance for prompt design.

\textbf{LLM Consistency and Reliability.}
A growing number of studies such as PromptRobust \citep{zhu2023promptrobust} and Sclar et al. \cite{sclar2024quantifying} have investigated prompt robustness and showed that contemporary LLMs are very sensitive to semantically equivalent promptings with subtle changes. This raises the question of reliability of LLMs obtained from single evaluations in few-shot settings. Wang et al. \cite{wang2023selfconsistency} develop a self-consistency decoding strategy to improve accuracy on arithmetic and common reasoning tasks. While it works effectively for structured reasoning, this approach is not deliberately designed for open-ended data science problems, whose outcomes may span numerical results, reasoning texts, code, and charts. \citep{wadi-fredette-2025-monte} uses a Monte Carlo based method to explore the reliability of LLM cognitive bias in social sciences and business.

Most of the existing studies (as summarized in Table \ref{tab:comparison}) do not explore the distribution of LLM behaviour and responses over repeated trials. To address this gap, we adapt Monte Carlo (MC) methods from classical statistics to LLM evaluation. By repeating each prompt about 100 times under the same conditions, we obtain empirical response distributions of task completion, accuracy, and code executability, which reveal systematic instability patterns that single-pass evaluation misses entirely.


\section{The \textsc{DS-QA} Benchmark}
\label{sec:ds-qa}
\begin{figure*}[!ht]
\centering
\includegraphics[width=.9\linewidth]{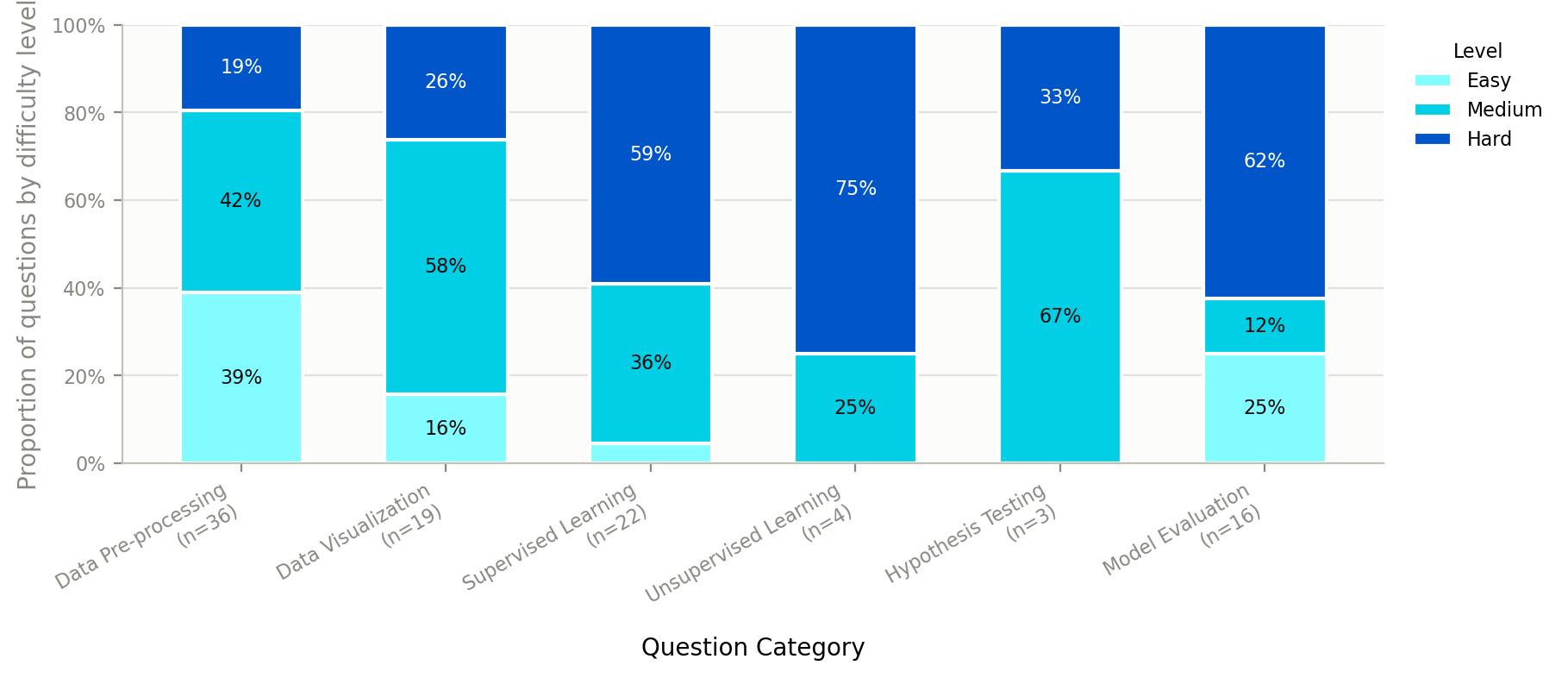}
\caption{Difficulty level distribution per macro question category.}
\label{fig:conceptvsdifficulty}

\end{figure*}
The proposed \textsc{DS-QA} benchmark consists of 100 data science question-answer pairs. The \textsc{DS-QA} pairs were from different undergraduate level statistical modules at the University of Edinburgh, including Introduction to Data Science (IDS) and Statistics Year 2 (StatY2). All these questions are collected in a jsonl file format. 

Our curated questions extend the four data science quadrants mentioned earlier by \cite{DeBie2022} to six specific question concepts, including:
\begin{itemize}
    \item Data Understanding and Wrangling
    \item Data Cleaning, Pre-processing and Transformations
    \item Data Summary Statistics and Interpretations
    \item Data visualisation - EDA
    \item Statistical Models
    \begin{itemize}
\item Supervised Learning (Linear Regression modeling, Variations of regression (i.e. Lasso or Ridge), 
Classification models such as Logistic Regression model, 
Tree based model, Model comparison type questions, Bayesian Regression)
\item Unsupervised Learning (i.e. Principal Component Analysis (PCA), Clustering methods)
    \end{itemize}
\item Statistical hypothesis testing-related questions
\end{itemize}

Beyond these macro question concepts, we further define three difficulty levels (easy, medium vs. hard), which are categorised by clarity and complexity. The specific definitions are:
\begin{itemize}
\item A question is classified as \textbf{easy} if it can be quickly read from the dataset with only one or a few computational steps. 
\item A question is classified as \textbf{medium} if it may require general reasoning and multiple computational steps. 
\item A \textbf{hard} question requires strong logical reasoning and complicated computational steps.  
\end{itemize}
Overall, the created questions range on the above list and vary by easy, medium vs. hard difficulty levels. Figure \ref{fig:conceptvsdifficulty} gives an overview of question categories and difficulty levels. Most \textsc{DS-QA} questions are associated with a specific dataset and the benchmark contains 16 datasets in total. Further details of task types, question IDs and their associated difficulty levels can be found in Appendix \ref{appendix.dataconstruct}.

\section{Evaluation Framework}
\label{sec:evalframework}
\begin{figure*}[!ht]
\centering
\includegraphics[width=\linewidth]{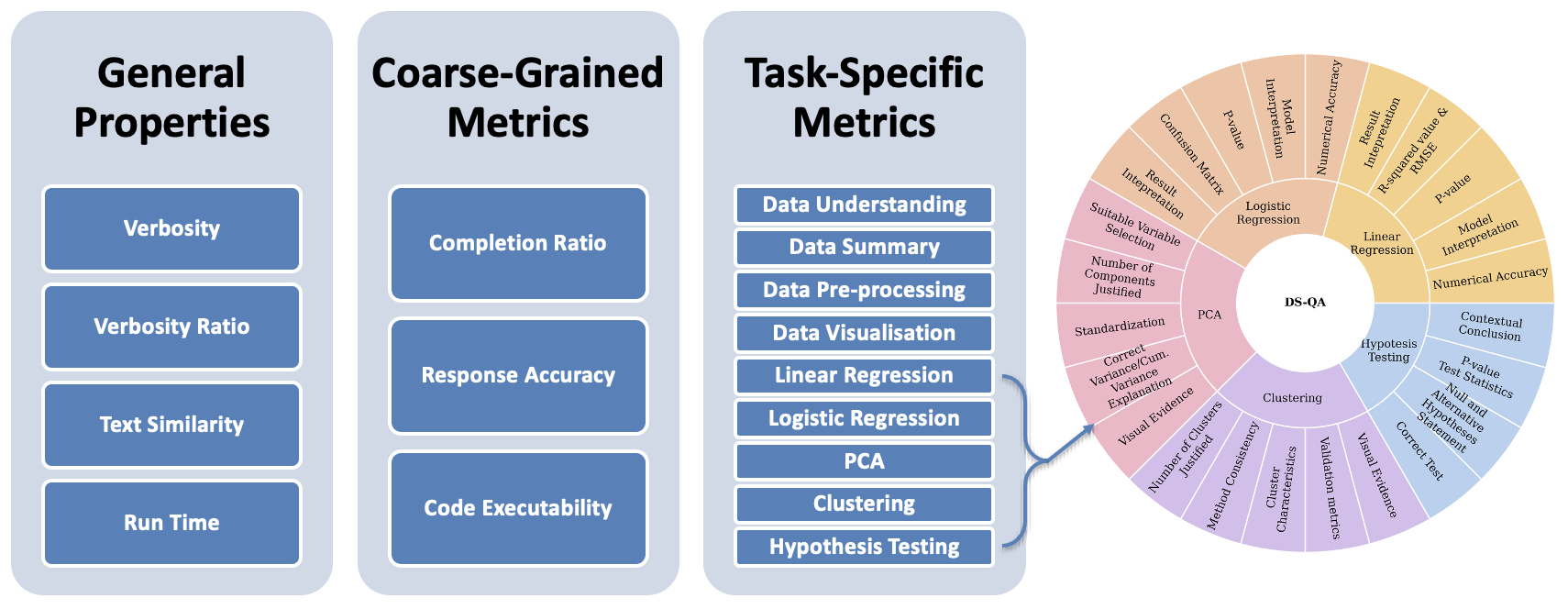}
\caption{A summary of the proposed multilayer evaluation framework, including general, coarse-grained and task-specific metrics. Regarding task-specific metrics, we introduce 5 functional task types on the basis of 4 elementary task types, which are Data Understanding, Data Summary, Data Pre-processing, and Data Visualisation. The pie chart on the right shows a summary of 5 functional task types and their task-specific metrics.} 
\label{fig:chart_evaluate}

\end{figure*}

Figure \ref{fig:chart_evaluate} gives a summary of our multilayer evaluation framework.

\textbf{General Properties.} We define four general properties (Verbosity, Verbosity Ratio, Text Similarity, and Run Time) to measure LLMs. 
\begin{itemize}
    \item \textbf{Verbosity}: General length for the number of tokens or words to count the verbosity of the generated text response in general except the coding component. 

    \item \textbf{Verbosity Ratio (input / output)}: It is the ratio of number of input tokens (words) and number of output tokens (words).

    \item \textbf{Text Similarity}: Similarity measures such as Jaccard Index for the text part of the generated response to compare with the first response as reference or ground truth for some tasks, when it is applicable. 
    
    \item \textbf{Run Time}: It measures the total runtime of the response creation per iteration. 

\end{itemize}

\textbf{Coarse-Grained Metrics.} In addition to general properties, we further define three coarse-grained metrics (Completion Ratio, Response Accuracy, and Code Executability). One of the given items below (Response Accuracy) will be combined later with the task specific cases to cover the nuances in generated responses across multiple promptings.
\begin{itemize}
    \item \textbf{Completion Ratio (CR)}: It is equal to 0 if the executed thread is failed/not completed, or 1 if the executed thread completed with an acceptable outcome (either matching with ground truth or not). 
    
    \item \textbf{Response Accuracy (RA)}: It takes either 0 or 1 if the reported result is matching with the ground truth (GT) or not, The GT can be numeric, string, or vector. If the output is decimal value, or if the match appears partially, the output will be automatically rounded to the closest value.  
    
    \item \textbf{Code Executability}: This tests whether the generated code is directly executable in a different environment or not, either taking 2 (it is executable and has solved the problem successfully), taking 1 (it is available but not executable directly) or 0 (no code).

\end{itemize}

\textbf{Task-Specific Metrics.} In this level, we look into 9 popular task types, which are Data Understanding, Data Summary, Data Pre-processing, Data visualisation, Linear Regression, Logistic Regression, Principal Component Analysis, Clustering, and Hypothesis testing. The set of criteria for Data visualisation is defined below. For the other task types, a varying set of criteria can be found in Appendix \ref{appendix.evaluate_taskspecific}. The task-specific metrics for specific questions are designed not only to focus on a single numerical answer. For the sake of simplicity, out of 5, each response can be evaluated based on the created metrics individually and repeatedly using binary marking.

\subsubsection*{Data visualisation}
\begin{itemize}
    \item \textbf{Data visualisation completeness}: It equals 1 if the requested data visualisation is generated, otherwise 0 instead. If this takes 0 there is nothing further to compute indeed.
    \item \textbf{Aesthetic mapping}: Each data variable is unambiguously mapped to an appropriate aesthetic (e.g., a continuous variable to position, a categorical variable to color), and the mappings are clearly documented in a legend or caption.
    \item \textbf{Geometric objects/layers}: The chosen geometric objects and functions (e.g., geom\_bar() for counts, geom\_point() for scatter) match the data structure and analytical intent given by the question, with no misuse or unsuitable geometric selection
    \item \textbf{Scales and coordinate system}: Axes use appropriate, non‑truncated scales with meaningful breaks; and any coordinate transform (e.g., log scale, coord\_flip()) is applied suitably, and all components are fully readable.
    \item \textbf{Visualisation-interpretation matching}: The correct use of words and abbreviations in the created plot and its associated interpretations.
\end{itemize}

\textbf{Final Score.} To quantify the numerical differences between LLM-generated responses across different task types and difficulty levels, we use the following formula:
\begin{align}
\text{Score} &= \text{Completion Ratio} +\text{Response Accuracy} \nonumber \\
    &+ \text{Code Executability}.
\end{align}
Recall that both Completion Ratio and Response Accuracy metrics return binary values. We normalize the Code Executability metric to be within $[0,1]$ and hence a score ranges from $0$ to $3$.

To measure the instability of LLM-generated responses for each question and each difficulty level, we take the standard deviation of the final scores into account; that is:
\begin{equation}\label{eq:inst} \text{Instability} (\sigma) = \sqrt{\frac{\sum_{i} (\text{Score}_i - \text{Mean\_Score})^2}{N}}
\end{equation}
Let $\bar{\sigma}\in [0,1]$ denote the normalized standard deviation of scores. The numerical difference is considered as significant if $\bar{\sigma} \ge 0.3$.

\section{Results}
\label{sec:results}
In this study, we evaluated a subset of 41 questions that were selected to cover all task categories. All the results and analyses reported below are based on this subset.

This section will look into the overall instability score distribution, a correlation heatmap of evaluation metrics, the statistical relationship between token verbosity ratios and word verbosity ratios, and initial insights with regard to different OpenAI models and temperatures.

\begin{figure}[!ht]
\centering
\includegraphics[width=.7\linewidth]{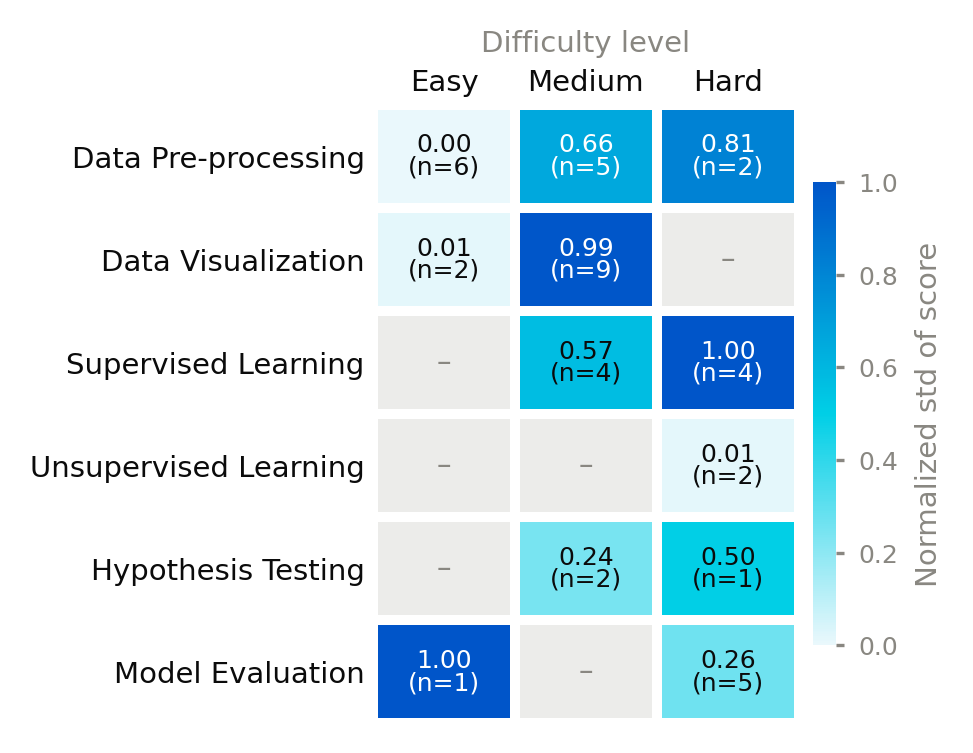}
\caption{Instability distribution by category and difficulty level for \textsc{gpt-4o} at a temperature of 1 (on the same 41-question subset with 100 runs per question).}
\label{fig:result.overall_std}

\end{figure}
\subsubsection*{\textbf{Instability Score Distribution.}} 
Figure \ref{fig:result.overall_std} summarises the distribution of score standard deviation across 100 iterations for \textsc{gpt-4o}. It is noticeable that for each task type except Model Evaluation, the instability increases with a higher difficulty level. Regarding Data Pre-processing, \textsc{gpt-4o} performs stably in easy-level tasks ($0\%$), but its performance is highly volatile in medium tasks ($66\%$) and hard tasks ($81\%$). Moreover, \textsc{gpt-4o} is highly unstable in medium and hard difficulty levels, especially involving the task types of Data Pre-processing, Data Visualisation, and Supervised Learning. The medium-level Data Visualisation tasks reach an instability of $0.99$, while the hard-level Supervised Learning tasks even reach $1.00$. These imply that repeated runs of an identical question return fully correct and failed answers almost evenly, and hence a single run on these questions is close to a coin toss. One illustrative example for Data Visualisation will be discussed later in Section \ref{sec:inconsist_example} and more examples can be found in Appendix \ref{appendix.examples}. 

\subsubsection*{\textbf{Metric Correlation.}} 
\begin{figure}[!ht]
\centering
\includegraphics[width=.7\linewidth]{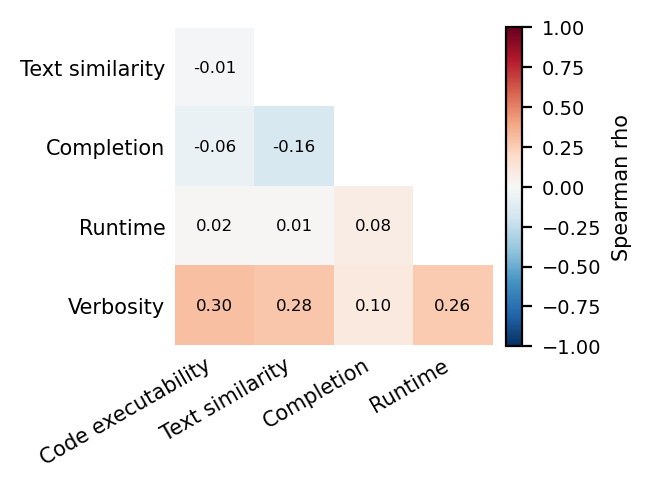}
\caption{Correlation heatmap for reliability indices, including both general properties and coarse-grained metrics.}
\label{fig:result.correlation}

\end{figure}

Figure \ref{fig:result.correlation} presents the correlation heatmap across metrics from the first two layers of our evaluation benchmark in Section \ref{sec:evalframework}. The evaluation metrics generally exhibit weak pairwise correlations, suggesting that they may capture complementary aspects of model performance. Verbosity shows weak positive correlations with code executability, text similarity, and runtime, with coefficients no greater than 0.30. This can be interpreted as a natural correlation: responses with a higher level of verbosity take longer runtime and possibly lead to more logical statements, which support more accurate code. Overall, the designed metrics from our evaluation benchmark are not redundant and capture multiple sources of instability. 

\vspace{-.3cm}
\subsubsection*{\textbf{Verbosity Analysis.}}
\label{subsec:result.verbosity}
Previous studies on verbosity have measured verbosity by default word count. However, for complicated data science tasks, token verbosity ratio is not equivalent to word verbosity ratio, even with similar trend.

\begin{figure}[!ht]
\centering
\includegraphics[width=\linewidth]{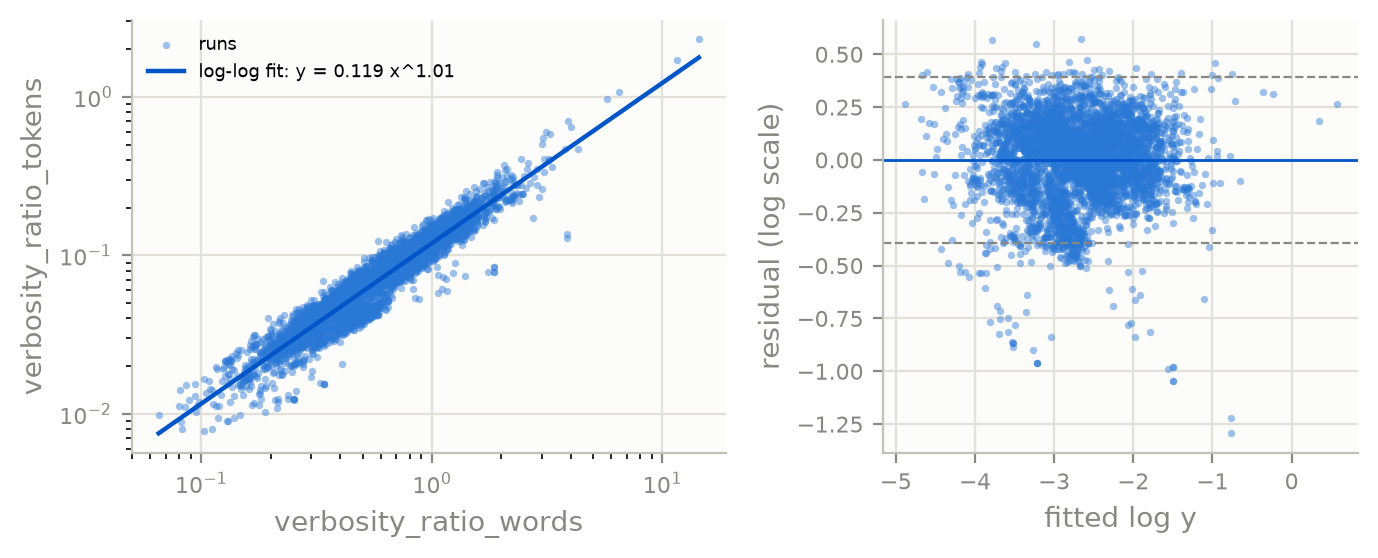}
\caption{Relationship between token verbosity ratio and word verbosity ratio for \textsc{ gpt-4o} at a temperature of 1. The left chart visualises the log-log model fitting and the right chart shows the residual plot with $4.5\%$ points beyond a $\pm 2$ SD interval.}
\label{fig:result.verbosity}

\end{figure}

Let $x$ denote word verbosity ratios and $y$ for token verbosity ratios. Figure \ref{fig:result.verbosity} (left) plots token verbosity ratios against word verbosity ratios on log-log axes and we fit the model with the simulation outcomes of $45$ questions with $100$ iterations per question. The Spearman coefficient is 0.953 and the Pearson correlation is 0.956. The points fall close to a straight line in log–log space, indicating a power-law relationship. The fitted exponent is 1.01 with a 95\% Confidence Interval [0.95, 1.07], which is statistically indistinguishable from 1. Therefore, it is consistent with the raw-scale linear fit and the fitted relationship is: 
\begin{align*}
\text{Token Verbosity Ratio} = -0.01 + 0.13\times\text{Word Verbosity Ratio}
\end{align*}
For the residual plot, most points ($> 95\%$) fall within the $\pm 2$ standard deviation (SD) interval, implying that the residuals have a roughly constant spread in log space. Moreover, the results still hold if we extend the single-model setting (\textsc{gpt-4o}) to a set of models (\textsc{gpt-4o},\textsc{gpt-4.1-mini},\textsc{gpt-4.1-nano}).

\subsubsection*{\textbf{Different models.}}
\begin{figure}[!ht]
\centering
\includegraphics[width=.8\linewidth]{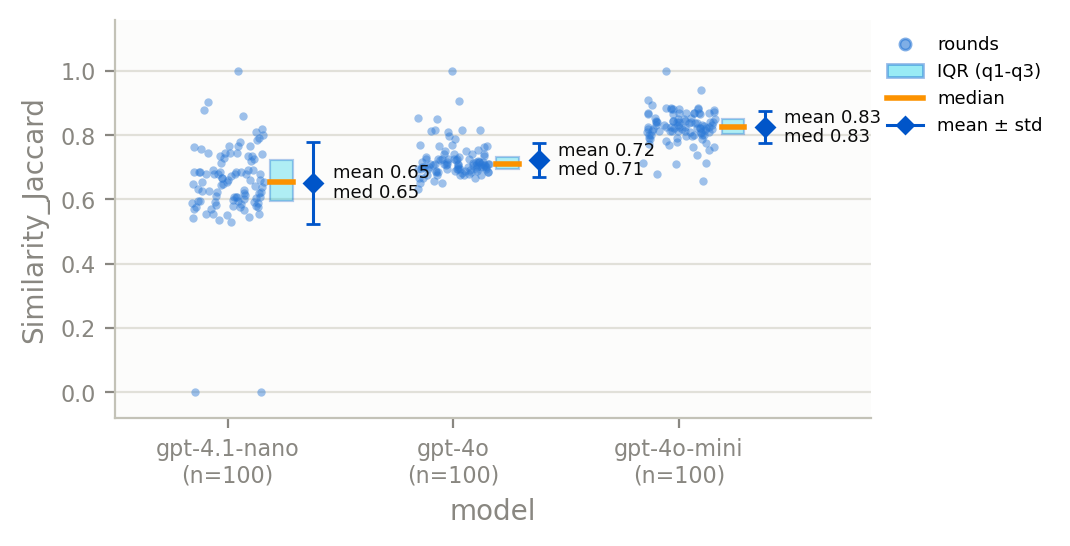}
\caption{Text similarity distribution by model with a temperature of 1 on id=56.}
\label{fig:result.model}

\end{figure}

Figure \ref{fig:result.model} shows the text similarity distribution of \textsc{gpt-4o}, \textsc{gpt-4o-mini}, and \textsc{gpt-4.1-nano} at a temperature of 1. The model \textsc{gpt-4o-mini} has the highest average Jaccard similarity level (0.83), followed by \textsc{gpt-4o} (0.72) and \textsc{gpt-4.1-nano} (0.65). For this question, \textsc{gpt-4o} shows the most divergent token usage distribution ranging from 480 to 1500, while the other models \textsc{gpt-4o-mini} and \textsc{gpt-4.1-nano} converge to consume about 500 and 650 tokens, respectively. It is surprising that \textsc{gpt-4.1-nano} with similar tokens returns responses with moderately volatile levels of text similarity, while \textsc{gpt-4o} with significantly different tokens tends to produce responses with a higher level of text similarity on average.  

\subsubsection*{\textbf{Different Temperatures.}}
\begin{figure}[!ht]
\centering
\includegraphics[width=.8\linewidth]{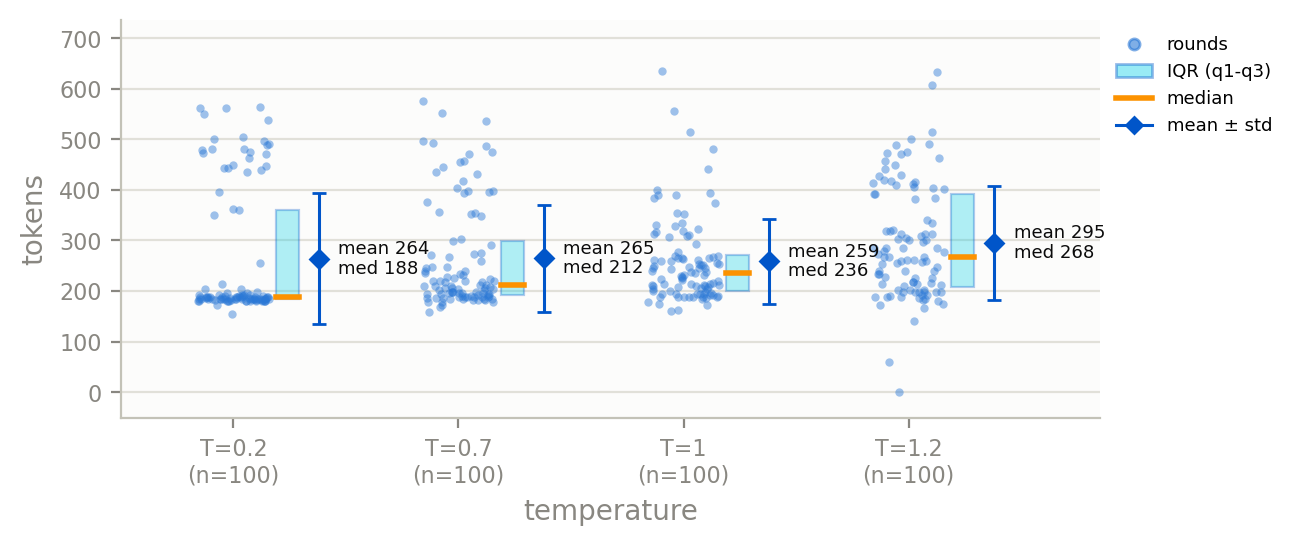}
\includegraphics[width=.8\linewidth]{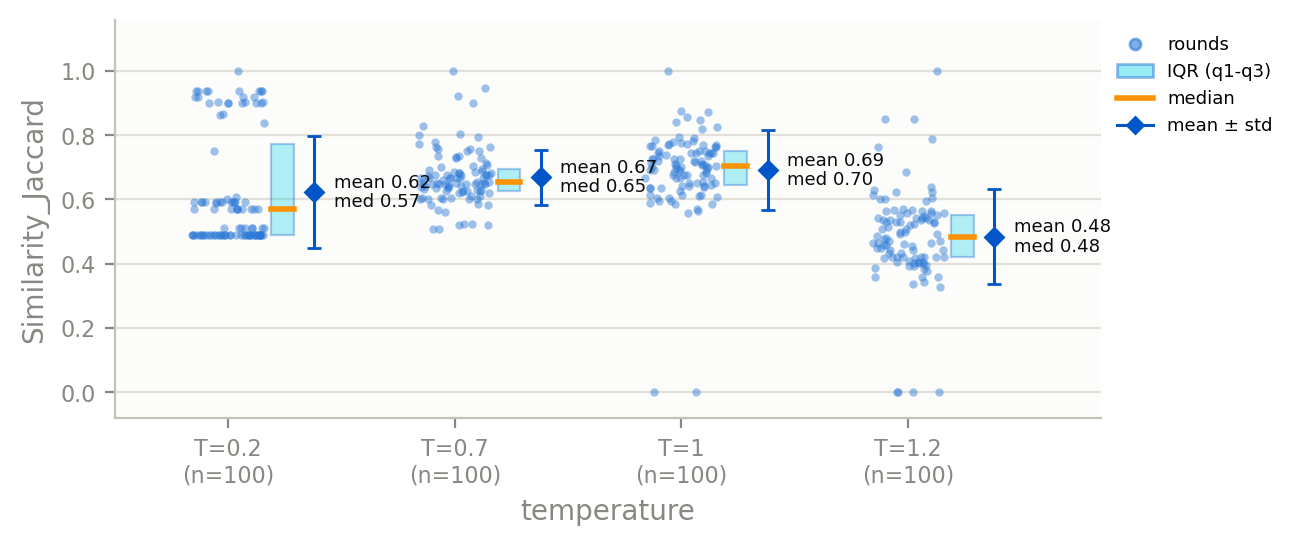}
\caption{Token usage distribution (top) and text similarity distribution (bottom) by temperature for \textsc{gpt-4.1-mini} on id=21.}
\label{fig:result.temp}

\end{figure}

Figure \ref{fig:result.temp} shows the performance of \textsc{gpt-4.1-mini} at a set of temperatures (0.2, 0.7, 1.0, and 1.2). Regarding token usage distribution, the average token remains between 259 and 265 when $T \le 1.0$ and steeply increases to $295$ at a temperature of 1.2. When $T=0.2$, there are clearly two clusters and the model uses 180-200 tokens in many iterations. The two clusters gradually converge from $T=0.2$ to $T=1.0$. When $T=1.0$ the token values are relatively evenly distributed within 170-400. The widest distribution occurs at $T=1.2$, continuous from 140 to 630. One interpretation is that a higher temperature ($T>1$) brings more possibilities of random changes.

A similar phenomenon can be found in the chart of text similarity. The value of text similarity becomes larger when text responses are more similar with less random changes. When T =1.2, the widest distribution corresponds to the lowest text similarity. From T = 0.2 to T = 1.0, the values tend to converge and have a higher level of text similarity.

\vspace{-.3cm}
\section{Illustrative example}
\label{sec:inconsist_example}
Here we provide one inconsistency example from the most unstable categories in the sense of \eqref{eq:inst}, i.e. Data Visualisation. 

Consider a hard-level example for the \textit{evals.csv} dataset: 

\vspace{0.2cm}
\textit{"Visualize the distribution of score in the dataframe evals. Is the distribution skewed? What does that tell you about how students rate courses? Is this what you expected to see? Why, or why not?"}
\vspace{0.2cm}

\begin{figure}[htbp]
\centering
\begin{subfigure}[t]{0.45\textwidth}
\centering
\includegraphics[width=.88\linewidth,height=3.5cm]{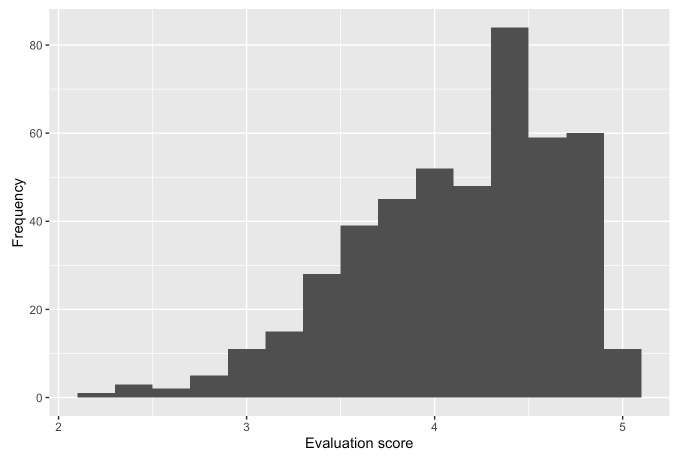}
\caption{GT Plot}
\label{fig:Qid21_groundTruth}

\end{subfigure}
  \hfill
\begin{subfigure}[t]{0.45\textwidth}
    \centering
\includegraphics[width=.85\linewidth,height=3.5cm]{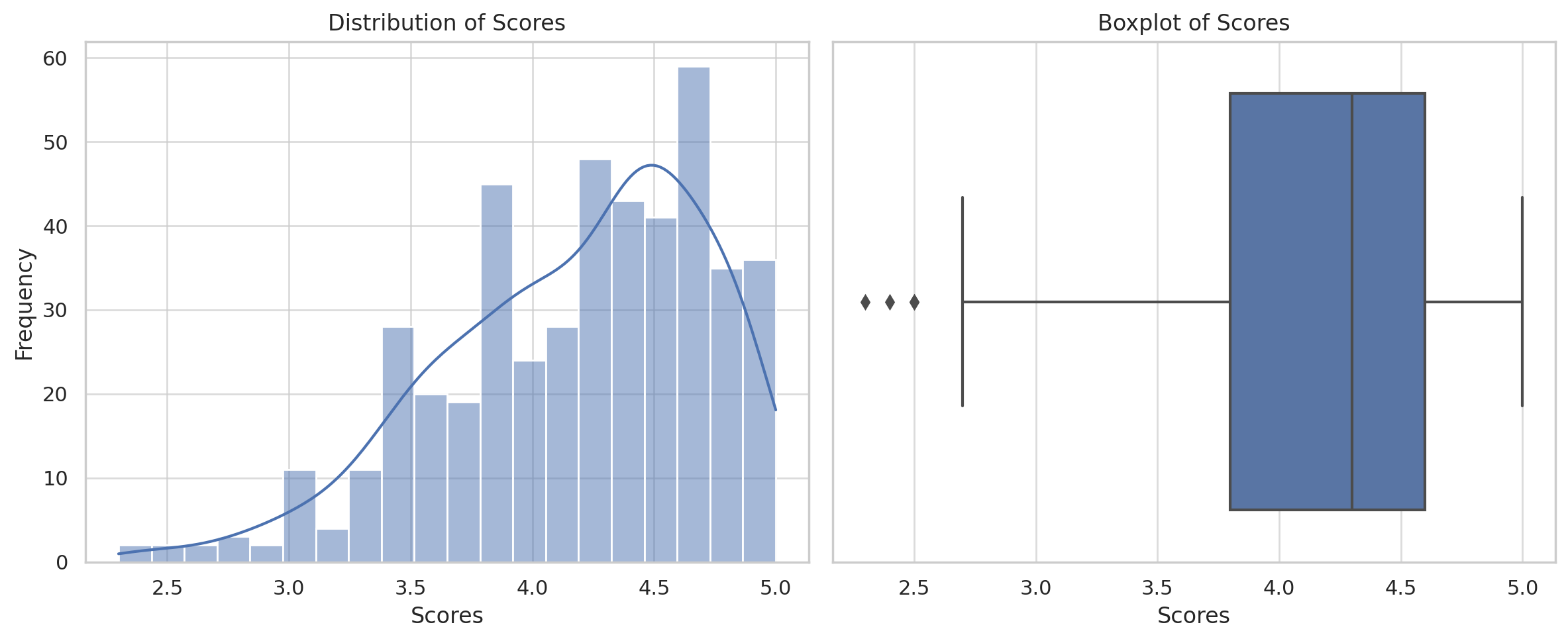}
    \caption{Output at round 22}
    \label{fig:Qid21_round22}
    
  \end{subfigure}
  \hfill
\begin{subfigure}[t]{0.45\textwidth}
    \centering
\includegraphics[width=.88\linewidth,height=3.5cm]{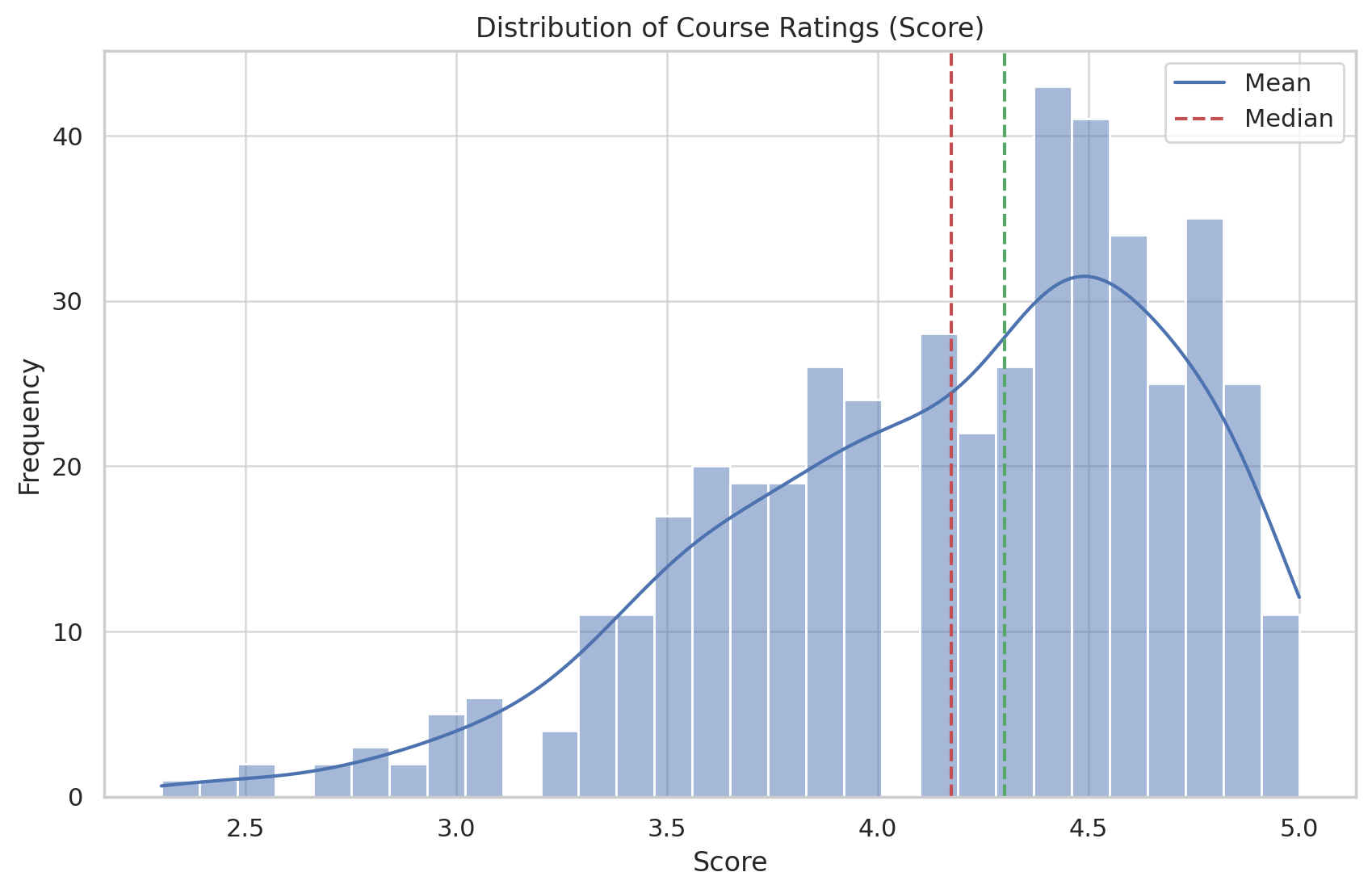}
    \caption{Output at round 41}
    \label{fig:Qid21_round41}
    
  \end{subfigure}
\hfill
\begin{subfigure}[t]{0.45\textwidth}
\centering
\includegraphics[width=\linewidth,height=4cm]{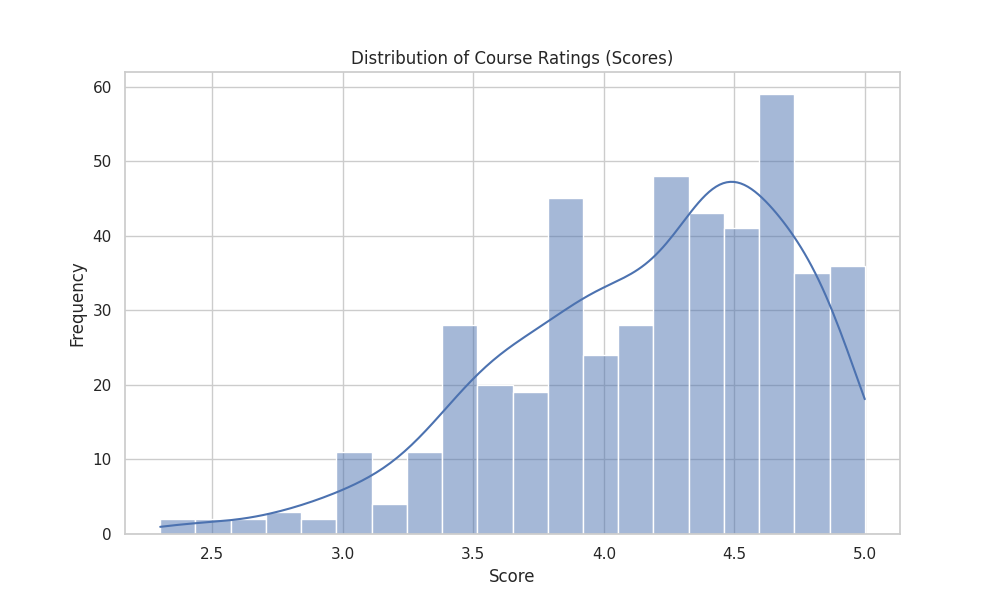}
\caption{Output at round 51}
\label{fig:Qid21_round51}

\end{subfigure}
\caption{An illustrative example for instability in Data Visualisation, including the Ground Truth chart and three generated images for question id=21 over the \textit{evals} dataset.}
\label{fig:Qid21_visuals_comparison}
\end{figure}

The average score of aesthetic mapping is $87\%$ and the main reason for losing this point is that frequency information is not used for y-axis as anticipated. Additionally, due to similar inconsistencies or incorrect usage, the average scores of geometric object and visualisation-Interpretation matching are $76\%$ and $68\%$, respectively. A few unexpected results emerge within 100 iterations, including the addition of a boxplot (see Figure \ref{fig:Qid21_round22}) and the occurrence of odd gaps in the created histograms (see Figure \ref{fig:Qid21_round41}). As identified from Figure \ref{fig:Qid21_visuals_comparison}, the Scales \& Coordinate System metric has literally had \textbf{zero success} because the values are somehow trimmed in the created histogram, meaning it never matches the GT image in our comparison.


\section{Limitation}
\label{sec:limit}
Our work focuses on data science tasks in English, and there is scope to further enrich the types and difficulty levels of data science tasks. Our evaluation covers 41 of the 100 benchmark questions and extending it to the full benchmark is left for future work. The experiments were run in the period from March to June 2025 on four OpenAI models (\textsc{gpt-4o},\textsc{gpt-4o-mini},\textsc{gpt-4.1-mini},\textsc{gpt-4.1-nano}), which have since been succeeded by the newer frontier models. We do not claim that the instability findings we report transfer to those models. Our protocol is designed for answering whether newer models are more stable. The central finding is that a single run and low dimensional evaluation criteria are unreliable bases for evaluating LLMs, which does not depend on a particular model. The next step is to apply the proposed protocol to the latest frontier models from several providers. 

\section{Conclusion}
\label{sec:conclu}
In this paper, we introduce a three-layer framework for evaluating LLM performance on data science tasks across a wide range of topics and three difficulty levels. The correlation heatmap also validates that the designed metrics can capture multiple sources of instability. Furthermore, our results show that the tested LLMs often fail to produce the elements identified as required by our evaluation framework. This suggests that the framework’s criteria could benefit prompt engineering directly, for example by explicitly requesting these elements in the prompt.

To study the reliability of LLMs over repeated runs, we propose a Monte Carlo assessment and our experimental results reveal substantial variability in data visualisation and hypothesis testing tasks, particularly at medium and hard difficulty levels, highlighting the importance of assessing consistency across repeated runs alongside response quality. Data Visualisation (medium-level) and Supervised Learning (hard-level) have an instability level of no less than $0.99$.

Moreover, we conduct a verbosity analysis and find that the token verbosity ratio increases by approximately 0.013 for every 0.1 increase in the word verbosity ratio, which can help measure LLM redundancy from the perspective of API costs. Regarding model comparison, we find that \textsc{gpt-4o-mini} has the highest level of text similarity with reasonable token usage, compared to \textsc{gpt-4o} and \textsc{gpt-4.1-nano}. With regard to different temperatures, the analysis of \textsc{gpt-4.1-mini} responses shows that a temperature larger than 1 is more likely to bring random changes to responses.

Some comments on future work are in order. First, we can further examine how sampling temperature affects both the accuracy and variability of responses. Second, it would be natural to use the proposed framework for comparing performance across a broad range of frontier LLMs. Finally, such a comparison would raise the question of which frontier model performs best and how strong the statistical evidence for its superiority is, given that it could simply have performed better by chance on the particular tasks considered. We leave such open problems for future analysis.

\section{Acknowledgements}
\label{sec:Acknowledgments}
This work was supported by the Generative AI Lab (GAIL) Seed Fund at the University of Edinburgh during the 2024-2025 round, for the period of January-June 2025.

\section*{Ethics and Privacy Statement}
Our work does not involve human participants and no personal data. The benchmark questions are adapted from university statistics teaching materials and all model outputs were obtained through commercial API under the providers’ terms of service. The main societal risk associated with this study is user over-reliance on answers from a Web-served LLM. Our work is intended to reduce this risk by making multi-round instability visible and measurable. Repeating each prompt 100 times consumes compute power and energy costs, which we consider justified for an evaluation study, but practitioners should consider how many repetitions are required for practical use. 

\bibliographystyle{ACM-Reference-Format}
\bibliography{references}

\newpage
\appendix
\section{\textsc{DS-QA} Data Construction}
\label{appendix.dataconstruct}
\subsubsection*{Format of Question-Answer pairs}
One example question is 
\vspace{0.2cm}
\begin{Verbatim}
{"id": 0, 
"question": "How many tourist attractions are there in the data set?", 
"concepts": ["Data understanding"], 
"format": "@count_visitor_value[count_visitor_value] 
where \"count_visitor_value\" is given in integer format.", 
"file_name": "UK-visitor-numbers.csv",  "level": "easy"}
\end{Verbatim}
\vspace{0.2cm}
and the corresponding answer is 
\vspace{0.2cm}
\begin{Verbatim}
{"id": 0, 
"question": "How many tourist attractions are there in the data set?", 
"concepts": ["Data understanding"], 
"format": "@count_visitor_value[count_visitor_value] where \"count_visitor_value\" is given in integer format.",
"file_name": "UK-visitor-numbers.csv", "level": "easy",
"common_answers": [["count_visitor_value", "348"]]}  
\end{Verbatim}
\vspace{0.2cm}
All the question-answer pairs can be found in the shared repository. See Table \ref{tab:question_summary} for a summary of task types and their corresponding question IDs at different difficulty levels. 

\begin{table*}[!ht]
\centering
\begin{tabular}{lp{0.22\linewidth}p{0.22\linewidth}p{0.22\linewidth}r}
\toprule
Task type & Easy & Medium & Hard & Total \\
\midrule
Data Understanding 
& 11.0, 31.0, 41.0 (7) 
& 24.0, 44.0, 66.0 (5) 
& (2) 
& 14 \\
Data Summary 
& 10.0, 32.0, 42.0, 46.0, 54.0 (5) 
& 12.0, 13.0, 16.1, 18.0 (6) 
& 17.0, 19.0, 20.0, 33.0 (4) 
& 15 \\
Data Pre-processing 
& 45.0 (2) 
& 15.1, 16.0 (2) 
& 37.0 (1) 
& 5 \\
Data visualisation 
& 14.0, 55.0 (3) 
& 14.1, 14.2, 22.0,  34.0, 47.0, 47.1, 53.0, 70.0 (11) 
& 15.0, 34.1, 35.0, 36.0, 39.1 (5) 
& 19 \\
Linear Regression 
& 68.0 (1) 
& 25.0, 38.0, 61.0, 62.0 (5) 
& 38.1, 48.0, 48.1, 49.0 (7) & 13 \\
Logistic Regression 
& -- & 69.4 (3) 
& 39.0, 69.0, 69.1, 69.2, 69.3 (7) & 10 \\
PCA & -- 
& 67.0 (1) & (1) & 2 \\
Clustering & -- & -- & 67.1 (2) & 2 \\
Hypothesis Testing & -- & (2) & (2) & 4 \\
Statistical Models & 40.0, 52.0, 68.1 (4) & 43.0, 51.0, 72.0, 73.0 (4) & 50.0, 61.1, 61.2, 64.0, 65.2, 67.2, 68.2 (8) & 16 \\
\bottomrule
\end{tabular}
\caption{Summary of task types, question IDs and their associated difficulty levels.}
\label{tab:question_summary}
\end{table*}

\section{Definitions of Task-Specific Metrics}
\label{appendix.evaluate_taskspecific}
\subsubsection*{Data Summary}
\begin{itemize}
    \item \textbf{Ratio of Accuracy:} It will be automatically compared with GT values.
    \item \textbf{Data context recognition:} Clearly define each variable, its units of measurement, and data source to ground the summary in context.
    \item \textbf{Appropriateness of chosen statistics:} Were the most relevant statistics reported for the data type within the context of question (e.g. median for skewed data, mode for categorical)?
    \item \textbf{Accuracy of interpretation:} Do the interpretations logically follow from the statistics? (e.g. not saying data is "normally distributed" based on wrong evidence)
    \item \textbf{Clarity of interpretation:} Are the interpretations understandable, concise, and easy to follow, or is there any confusing word selection or terminology usage?
\end{itemize}

\subsubsection*{Linear Regression}
\begin{itemize}
    \item \textbf{Ratio of Accuracy:} (will be calculated automatically via comparing GT value)
    \item \textbf{Correct model coefficient interpretation:} Understanding the magnitude and direction of the relationship between independent and dependent variables
    \item \textbf{p-value recognition:} Recognizing statistical significance on the model coefficients by reporting of p-values for each coefficient against a pre-specified $\alpha$-level (commonly 0.05).
    \item \textbf{Evaluation Metric Reported and Interpreted:} Assessing the model's goodness of fit and the proportion of variance explained by the model via $R^{2}$, and/or supplemented by predictive error metrics such as RMSE or others
    \item \textbf{Interpretation of Results:} Interpreting the results within the context of the specific research question and the problem domain.
\end{itemize}

\subsubsection*{Logistic Regression}
\begin{itemize}
    \item \textbf{Ratio of Accuracy} (will be calculated automatically via comparing GT value)
    \item \textbf{Correct model coefficient interpretation:} Understanding the magnitude and direction of the relationship between independent and dependent variables
    \item \textbf{p-value recognition:} Recognizing statistical significance on the model coefficients by reporting of p-values for each coefficient against a pre-specified $\alpha$-level (commonly 0.05).
    \item \textbf{Confusion Matrix recognition:} The confusion matrix details and derive class-specific error rates
    \item \textbf{Interpretation of Results:} Interpreting the results within the context of the specific research question and the problem domain.
\end{itemize}

\subsubsection*{Principal Component Analysis (PCA)}
\begin{itemize}
    \item \textbf{Suitable Variable Selection:} Identifying the suitable type of variables in the data by selecting. 
    \item \textbf{Number of Components Justified:} Is the decision on how many PCs to keep justified correctly?
    \item \textbf{Standardization of Input Data:}  Was scaling recognised and applied where needed before PCA?
    \item \textbf{Correct Variance/Cum. Variance Explanation:}  Are explained variance / cumulative variance ratios reported and understood?
    \item \textbf{Visual evidences:} Is the decision on the number of PCs supported by certain visuals like loading, biplot or scree plot. 
\end{itemize}

\subsubsection*{Clustering}
\begin{itemize}
    \item \textbf{Number of clusters justified:} Is the final number of obtained clusters justified correctly?
    \item \textbf{The methods consistency:} Is the requested method applied properly with the specification of certain parameters
    \item \textbf{Cluster characteristics:} Are the obtained cluster characteristics such as cluster size, cluster profile mentioned clearly
    \item \textbf{Validation metrics:} Are the certain report metrics like silhouette scores, Davies-Bouldin index, or other relevant statistics are recognised
    \item \textbf{Visual evidence:}  Is the decision on the number of clusters supported by certain visuals like scatter plots, dendrograms. 
\end{itemize}

\subsubsection*{Hypothesis testing}
\begin{itemize}
    \item \textbf{Correct Test/Test name Chosen:} Recognition of the correct statistical testing / test name
    \item \textbf{Null and Alternative Hypotheses Clearly Stated:} Clearly and correctly stated hypothesis test components with correct notation
    \item \textbf{Accuracy on test statistics/p-value and/or Critical region}  Having the correct final decision based on the testing procedure
    \item \textbf{Correct Interpretation of test statistics/p-value and/or Critical region}  Having the correct final decision based on the testing procedure
    \item \textbf{Contextual Conclusion:} Interpreting the final decision within the context of the specific research question
\end{itemize}

\section{Additional Illustrative Examples}
\label{appendix.examples}
\subsubsection*{Hypothesis Testing.}
\label{subsec:inconsist_ht}
Consider a medium-level example with the \textit{aeroplane.txt} dataset: 

\vspace{0.2cm}
\textit{"For now consider only the data stored in the vector first. We assume that the data are independent observations from $N(\mu, \sigma^2)$ and wish to conduct the following hypothesis test: H0 : $\mu$ equals 10 vs H1 : $\mu$ not equal 10. Define a suitable test statistic and state its distribution, assuming that the null hypothesis is true. Calculate the associated observed test statistic. Conduct the hypothesis test and clearly state your conclusions."}
\vspace{0.2cm}

"Correct Test/Test name Chosen" has $100\%$ success, whereas there is less consistency on the items such as "H0 and H1 are clearly stated", "Accuracy on Test Stat / p-value / Critical Reg" and "Contextual Conclusion" having values of  $73\%$, $72\%$ and $76\%$, respectively. Even if for some cases the test numerical results are not matching with the GT, the final interpretation seems reasonable so we had "Correct Interpretation of  Test Stat / p-value / Critical Reg" as $82\%$ of the time over 100 times repeated prompting. There were some cases that the solution is incomplete (asking the user to continue), no specific numerical outcome or using the sample mean in the interpretation rather than the population mean, as one of the challenging misuse for learners in general. 

\subsubsection*{Linear Regression.}
\label{subsec:inconsist_linear}
Consider another hard-level example from \textit{evals.csv}, with an overall performance of 4.34 out of 5 scores: 

\vspace{0.2cm}
\textit{"Fit a linear model called score\_bty\_fit to predict average professor evaluation score from average beauty rating (bty\_avg).Print the regression output using tidy(). Based on the regression output, write down the linear model."}
\vspace{0.2cm}

Our results show that GPT models achieve large success over the sub-items of Accuracy on results - $92\%$, Correct model coefficient interpretation - $99\%$, Evaluation Metric reporting - $99\%$ and Interpretation of Results - $93\%$ beyond looking at accuracy only. However, the recognition of the p-value is missing in $51\%$ of 100 cases.

\subsubsection*{PCA}
\label{subsec:inconsist_pca}
One example with an overall performance of 2.92/5 is from the \textit{duke\_forest.xlsx} dataset and at the hard level: 

\vspace{0.2cm}
\textit{"Run PCA for this data, with all suitable variables. Discuss whether it is appropriate to run scaled or unscaled PCA for this data."}
\vspace{0.2cm}

The LLM-generated response rarely includes the justification of the selected number of components or the numerical explanation of the explained variance or cumulative variance. Even if the code structure seems following the correct recipe mostly, the generated text response is mostly limited. To illustrate, it is observed that "No of components justified" - $16\%$ and "Correct Variance/ Cum Var explanation" - $26\%$ lower values compared to other items such as detecting the reasonable subset of numerical variables to apply PCA ("Suitable Variable Selection" - $98\%$), necessary scaling for the variables have different ranges ("Scaling Recognition" - $99\%$) and even the specific supportive visualization in half of the cases ("Visual Evidence" - $53\%$). 

\end{document}